\documentclass[
]{ceurart}

\usepackage{listings}
\usepackage{multirow}
\begin{document}

%%
%% Rights management information.
%% CC-BY is default license.
\copyrightyear{2022}
\copyrightclause{Copyright for this paper by its authors.
  Use permitted under Creative Commons License Attribution 4.0
  International (CC BY 4.0).}

%%
%% This command is for the conference information
% \conference{Forum for Information Retrieval Evaluation, December 9-13, 2022, India}

\title{A Twitter BERT Approach for Offensive Language Detection in Marathi}

% \copyrightyear{2022}
% \copyrightclause{Copyright for this paper by its authors.
%   Use permitted under Creative Commons License Attribution 4.0
%   International (CC BY 4.0).}

%%
%% This command is for the conference information
\conference{Forum for Information Retrieval Evaluation, December 9--13, 2022, India}

% \tnotemark[1]
% \tnotetext[1]{You can use this document as the template for preparing your
%   publication. We recommend using the latest version of the ceurart style.}

% \author[1]{Anonymous Authors}[%
% ]
% \fnmark[1]

\author[1, 3]{Tanmay Chavan}[%
email=chavantanmay1402@gmail.com
]
\fnmark[1]

\author[1, 3]{Shantanu Patankar}[%
email=shantanupatankar2001@gmail.com
]
\fnmark[1]

\author[1, 3]{Aditya Kane}[%
email=adityakane1@gmail.com
]
\fnmark[1]

\author[1, 3]{Omkar Gokhale}[%
email=omkargokhale2001@gmail.com
]
\fnmark[1]

\author[2, 3]{Raviraj Joshi}[%
email=ravirajoshi@gmail.com
]
\fnmark[1]

\address[1]{Pune Institute of Computer Technology, Pune}
\address[2]{Indian Institute of Technology Madras, Chennai}
\address[3]{L3Cube, Pune}

% \author[3]{Ilaria Tiddi}[%
% orcid=0000-0001-7116-9338,
% email=i.tiddi@vu.nl,
% url=https://kmitd.github.io/ilaria/,
% ]
% \fnmark[1]
% \address[3]{Vrije Universiteit Amsterdam, De Boelelaan 1105, 1081 HV Amsterdam, The Netherlands}

% \author[4]{Manfred Jeusfeld}[%
% orcid=0000-0002-9421-8566,
% email=Manfred.Jeusfeld@acm.org,
% url=http://conceptbase.sourceforge.net/mjf/,
% ]
% \fnmark[1]
% \address[4]{University of Skövde, Högskolevägen 1, 541 28 Skövde, Sweden}

%% Footnotes

\fntext[1]{These authors contributed equally.}

\begin{abstract}
    Automated offensive language detection is essential in combating the spread of hate speech, particularly in social media. This paper describes our work on Offensive Language Identification in low resource Indic language Marathi. The problem is formulated as a text classification task to identify a tweet as offensive or non-offensive. We evaluate different mono-lingual and multi-lingual BERT models on this classification task, focusing on BERT models pre-trained with social media datasets. We compare the performance of MuRIL, MahaTweetBERT, MahaTweetBERT-Hateful, and MahaBERT on the HASOC 2022 test set. We also explore external data augmentation from other existing Marathi hate speech corpus HASOC 2021 and L3Cube-MahaHate. The MahaTweetBERT, a BERT model, pre-trained on Marathi tweets when fine-tuned on the combined dataset (HASOC 2021 + HASOC 2022 + MahaHate), outperforms all models with an F1 score of 98.43 on the HASOC 2022 test set. With this, we also provide a new state-of-the-art result on HASOC 2022 / MOLD v2 test set.
\end{abstract}


\maketitle

\section{Introduction}
Offensive speech detection in social media is a crucial task \cite{schmidt-wiegand-2017-survey}. The impact of cyberbullying and offensive social media content on society's mental health is still under research, but it is undeniably negative \cite{naslund2020social}. With the increasing number of social media users, offensive speech identification is a crucial task necessary to maintain harmony. In this work, we focus on offensive language detection in the low-resource Marathi language.

Marathi is an Indo-Aryan language predominantly spoken in the Indian state of Maharashtra. Marathi is a rich language derived from Sanskrit and has 42 dialects. Spoken by 83 million people, it is the third-largest spoken language in India and the tenth in the world. 

%Our team, Optimize\_Prime, participated in the HASOC 2022 \cite{hasoc2022mergeoverview} shared task on Offensive Language Identification in Marathi, Subtask-3A: Offensive Language Detection. The shared task consisted of classifying tweets as offensive or non-offensive. 

In this work, we explore different mono-lingual and multi-lingual pre-trained BERT transformer models for offensive speech detection. These models are evaluated on HASOC 2022 dataset. It is a binary offensive language identification dataset in Marathi. Since the evaluation data is based on social media, we focus on Marathi Twitter BERT models and show their superior performance. We evaluate the performance of MuRIL \cite{muril}, MahaTweetBERT, MahaTweetBERT-Hateful \cite{patankar2022spread} and MahaBERT \cite{joshi-2022-l3cube} on the HASOC-2022 or MOLD v2 \cite{Zampieri2022} test set. We further explore data augmentation from other existing hate speech detection datasets like HASOC 2021 \cite{hasoc2021overview}, HASOC 2022 \cite{hasoc2022overview}, and L3Cube-MahaHate \cite{patil-etal-2022-l3cube}. We show that the external datasets are helpful in further improving the performance of the model.  
We provide a detailed analysis of the performance of various mono-lingual and multi-lingual models on the two datasets and reflect on the shortcomings of the models.

The main contributions of this work are as follows:
\begin{itemize}
    \item We explore Marathi Tweet BERT models for the task of social media offensive language identification and show their superior performance. We specifically consider the recently released MahaTweetBERT and report new state-of-the-art results on HASOC 2022 or MOLD v2 test set. We show that pre-training on tweets data is indeed helpful on the downstream social media tasks.
    \item We show that hateful BERT models do not provide the best results although intuitively these BERT models should work well on the offensive language tasks. These observations were initially presented in \cite{patankar2022spread}, this work also re-iterates their observations.
    \item We show that different available hate speech data sources can be combined to further improve the performance. We present a data augmentation approach by considering two extra external datasets for training.  
\end{itemize}
To the best of our knowledge, this is the first work to explore the Twitter BERT model and external data augmentation using multiple data sources for the Marathi offensive language identification task.

\section{Related Work}
Since the advent of social media, detecting offensive language has become an imperative task. 
Earlier works like \citet{chen2012detecting} used a lexical analysis approach to detect hate speech. Later works like \citet{kumar2018benchmarking} present a more traditional machine learning-based approach for hate speech detection using feature engineering and models like support vector machines. \citet{aroyehun2018aggression} uses word embeddings from word2vec, Glove, SSWE, and fastText and uses seven deep learning models, including CNNs, LSTMs, and Bi-LSTMs to detect hate speech from Facebook posts. The introduction of the attention layer and transformers in \citet{vaswani2017attention} has led to the emergence of various transformer architectures like BERT \cite{devlin2018bert}. These transformers can be fine-tuned on downstream tasks like hate speech to yield exceptional results. There has been some research studying the effect of contextual or domain-specific pre-training. HateBERT \cite{caselli2020hatebert}, and FBERT \cite{sarkar2021fbert} are BERT models pre-trained on specially curated hate speech data. Both models obtain better results than merely fine-tuning vanilla BERT on a target hate speech dataset.

While much work is available in high-resource languages like English or German, offensive language detection in low-resource languages is relatively less explored. Social media apps' measures to reduce offensive language are generally limited to high-resource languages. \citet{velankar2022review} and \citet{gaikwad-etal-2021-cross} demonstrate several challenges and limitations faced while performing offensive language detection in Marathi. These limitations warrant a more in-depth study in this particular domain. Multilingual models like MuRIL\cite{khanuja2021muril} have been known to perform well on hate speech datasets. However, some recent works \cite{velankar2022mono} in Marathi show that Monolingual models like MahaBERT\cite{joshi-2022-l3cube} achieve better results than their multilingual counterparts. Their observations were in the context of hate speech detection as well as general text classification. A much larger corpus consisting of 25000 distinct tweets named L3Cube-MahaHate for hate speech identification for Marathi was proposed in \cite{patil2022l3cube}.
In our approach, we fine-tune MuRIL, MahaTweetBERT, MahaTweetBERT-Hateful, and MahaBERT on a combination of HASOC 2022 \cite{hasoc2022overview}, HASOC 2021 \cite{hasoc2021overview}, and MahaHate data. Both of which are offensive/hate language detection datasets in Marathi.

\begin{table}[h]
    \caption{Dataset Description for HASOC 2021, HASOC 2022, and MahaHate datasets}
    \label{tab:dataset_description_tables}
\begin{tabular}{|c|c|c|c|}
\hline
\textbf{Datasets}   & \textbf{Offensive} & \textbf{Non-offensive} & \textbf{Total} \\ \hline
\textbf{HASOC 2021} & 1,205              & 669                    & 1,874          \\ \hline
\textbf{HASOC 2022} & 2,034              & 1,062                  & 3,096          \\ \hline
\textbf{MahaHate 2-class} & 18,750              & 18,750                  & 37500          \\ \hline
\textbf{MahaHate 4-class subset} & 6,250              & 6,250                  & 12,500          \\ \hline
\end{tabular}
\end{table}

\section{Dataset}

We fine-tune our models on three sets of data. The first dataset consists of the HASOC 2022 training data. The second dataset is a combination of HASOC 2021 and HASOC 2022 data. The third set is a combination of HASOC 2021 + HASOC 2022 + MahaHate datasets.

\subsection{HASOC 2022}
The HASOC 2022 dataset consists of text from social media. The data points are labeled as offensive and not offensive. We use 70\% of the training data to train the model and 30\% of the data for validation. The data consists of 3096 data points. Out of these, 2034 are offensive, and 1062 are non-offensive. This dataset is identical to the MOLD v2 dataset, so we use the names "HASOC 22" and "MOLD v2" interchangeably throughout this paper.

\begin{figure}[h]
\begin{center}
    \includegraphics[width=10cm]{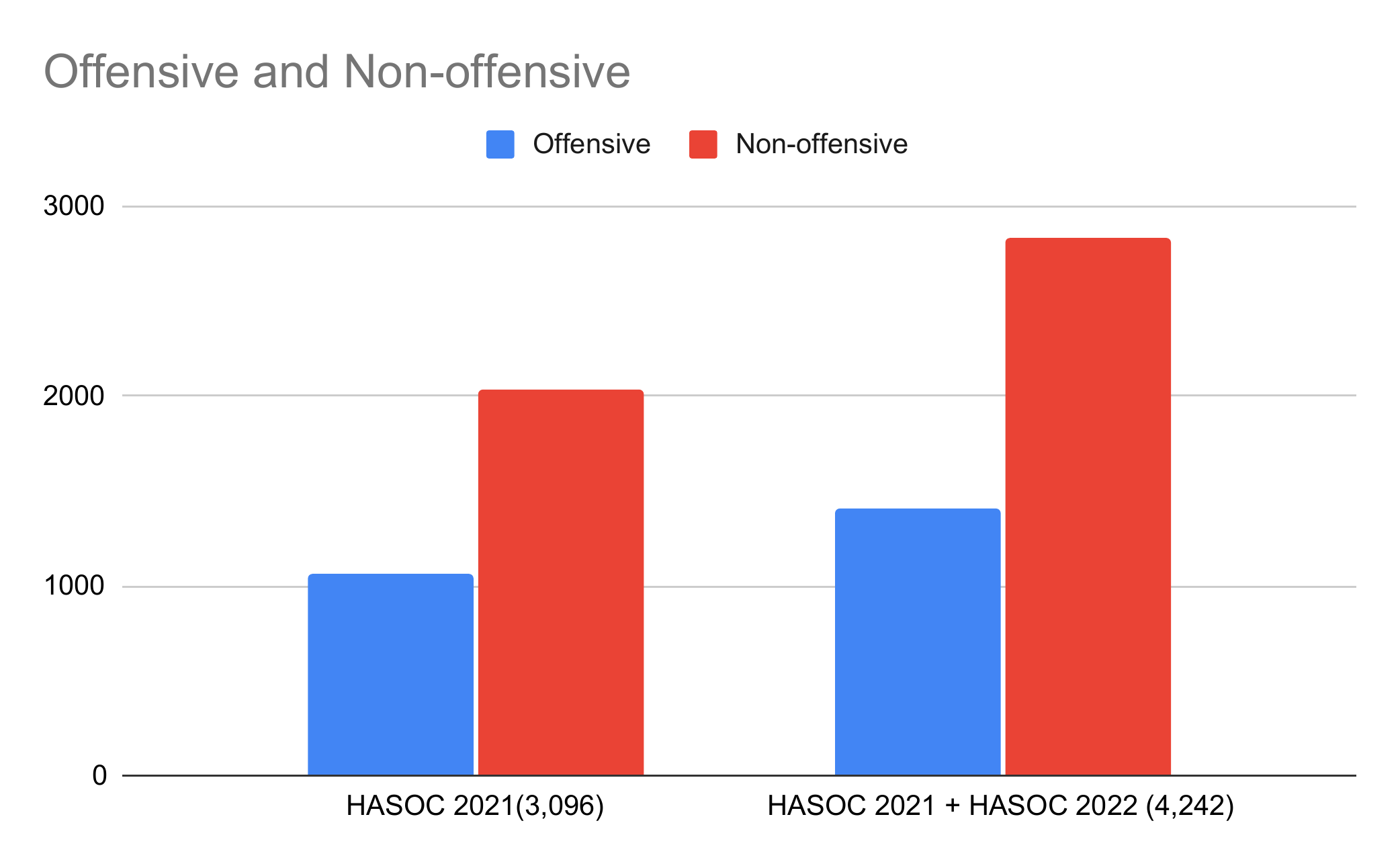}
\end{center}
\end{figure}

\subsection{HASOC 2021 + HASOC 2022}
The HASOC 2021 dataset consists of text obtained from Twitter. The tweets are categorized as offensive and non-offensive. There is a total of 1874 tweets in the dataset. Of these, 1205 are offensive, and 669 are non-offensive. We combine this with the HASOC 2022 data and use the combined dataset for fine-tuning.

\subsection{HASOC 2021 + HASOC 2022 + MahaHate}
The MahaHate dataset  \cite{patil-etal-2022-l3cube}  has four class and two class configurations. The four-class configuration consists of 6250 tweets each for hate, offensive, profane, and none labels. We combine hate and none categories with HASOC 2021 and HASOC 2022 datasets and use the combined dataset for fine-tuning (HASOC 21 + HASOC 22 + MahaHate 4c-subset). The two-class configuration contains 18,750 tweets each for hate and non-hate. We combine this dataset with HASOC 2021 and HASOC 2022 datasets and use the combined dataset for fine-tuning (HASOC 21 + HASOC 22 + MahaHate 2c).

\section{Experiments}
We experiment with several models in the course of our experiments. We choose to use BERT-based models as they have shown promising results for text classification. Pre-training these models on large datasets has proven to yield outstanding results on downstream classification tasks in the same language. We describe our methods below.

\subsection{Data Preprocessing}

We preprocessed the data to obtain better results on the classification task. Although the dataset was significantly clean, we performed cleaning operations to ensure the ideal conditions of the data. The provided dataset had redacted username mentions in the tweets and replaced them with the placeholder text '@USER' to protect the identity of the original author of the tweets. We chose to remove the placeholder texts. Our preprocessing methods also cleaned any newline hashtags, URLs, empty parentheses, and newline characters.

\subsection{Model training}

We use the HuggingFace \cite{wolf-etal-2020-transformers} framework for using the models. All the tweets were tokenized by the tokenizer specified by the model before being used by the model. The tokenized text was used by the model backbone. The model's output was then processed through a fully connected feed-forward network layer. We finally used softmax. The models used in this work are described below.

\begin{itemize}
    \item \textbf{MuRIL} is a BERT-based model pre-trained on a large multilingual dataset encompassing several Indian languages.
    \item \textbf{MahaTweetBERT} \footnote{MahaTweetBERT link: https://huggingface.co/l3cube-pune/marathi-tweets-bert} \cite{patankar2022spread} is a BERT-based model pre-trained on L3Cube-MahaTweetCorpus (848 million tokens), a large monolingual dataset containing tweets written in the Marathi language.
    \item \textbf{MahaTweetBERT-Hateful} \footnote{MahaTweetBERT-Hateful link: https://huggingface.co/l3cube-pune/marathi-tweets-bert-hateful} is a model which is trained exclusively on hateful Marathi tweets extracted from the L3Cube-MahaTweetCorpus (23.7  million tokens). 
    \item \textbf{MahaBERT} \footnote{MahaBERT link: https://huggingface.co/l3cube-pune/marathi-bert-v2} \cite{joshi-2022-l3cube} is a BERT model pre-trained on L3Cube-MahaCorpus, a large Marathi dataset containing 725 million tokens.
\end{itemize}
   All of these models are freely available on HuggingFace.

We have used two datasets for training our models. We used the HASOC 2022 dataset and another combined dataset containing samples from the HASOC 2022 dataset and the HASOC 2021 dataset. We performed a split on the HASOC 2022 training dataset to obtain the validation dataset. We used the metric of macro F1 score as implemented in the scikit-learn module. We can see that the MahaTweetBERT model seems to perform very well on both datasets.

We have used the AdamW optimizer with a learning rate of 1e-5 and batch size of 32. We trained the models for 25 epochs. The hyperparameter values and number of epochs remain the same across all the models to maintain consistency between different results.

% - all models that were used and their brief description
% - system desciption: LLM + linear layer
% - model evaluation on validation dataset, final submission has 2 models
% - all hyperparameters

\begin{table}[]
\caption{Results of all models trained on the HASOC-22 dataset and tested on the MOLD v2 test data}
\label{tab:ablation_results}
\begin{tabular}{|
>{\columncolor[HTML]{FFFFFF}}c |
>{\columncolor[HTML]{FFFFFF}}c |
>{\columncolor[HTML]{FFFFFF}}c |
>{\columncolor[HTML]{FFFFFF}}c |
>{\columncolor[HTML]{FFFFFF}}c |}
\hline
\textbf{Training Dataset}                                                                                                     & \textbf{Model}        & \textbf{Accuracy} & \textbf{F1 macro} & \textbf{F1 weighted} \\ \hline
\cellcolor[HTML]{FFFFFF}                                                                                                      & MuRIL                 & 90.58             & 90.58             & 90.58                \\ \cline{2-5} 
\cellcolor[HTML]{FFFFFF}                                                                                                      & MarathiBert V2        & 90.39             & 90.39             & 0.90.39              \\ \cline{2-5} 
\cellcolor[HTML]{FFFFFF}                                                                                                      & MahaTweetsBERT        & 91.56             & 91.56             & 91.56                \\ \cline{2-5} 
\multirow{-4}{*}{\cellcolor[HTML]{FFFFFF}HASOC 2022}                                                                          & MahaTweetsBERTHateful & 91.96             & 91.95             & 91.95                \\ \hline
\cellcolor[HTML]{FFFFFF}                                                                                                      & MuRIL                 & 96.47             & 96.47             & 96.47                \\ \cline{2-5} 
\cellcolor[HTML]{FFFFFF}                                                                                                      & MarathiBert V2        & 96.08             & 96.08             & 96.08                \\ \cline{2-5} 
\cellcolor[HTML]{FFFFFF}                                                                                                      & MahaTweetsBERT        & 96.08             & 96.08             & 96.08                \\ \cline{2-5} 
\multirow{-4}{*}{\cellcolor[HTML]{FFFFFF}HASOC 21 + HASOC 22}                                                                 & MahaTweetsBERTHateful & 95.88             & 95.88             & 95.88                \\ \hline
\cellcolor[HTML]{FFFFFF}                                                                                                      & MuRIL                 & 96.27             & 96.27             & 96.27                \\ \cline{2-5} 
\cellcolor[HTML]{FFFFFF}                                                                                                      & MarathiBert V2        & 97.06             & 97.06             & 97.06                \\ \cline{2-5} 
\cellcolor[HTML]{FFFFFF}                                                                                                      & MahaTweetsBERT        & \textbf{98.43}             & \textbf{98.43}             & \textbf{98.43}                \\ \cline{2-5} 
\multirow{-4}{*}{\cellcolor[HTML]{FFFFFF}\begin{tabular}[c]{@{}c@{}}HASOC 21 + HASOC 22 + \\ MahaHate 2c\end{tabular}}        & MahaTweetsBERTHateful & 97.45             & 97.45             & 97.45                \\ \hline
\cellcolor[HTML]{FFFFFF}                                                                                                      & MuRIL                 & 96.67             & 96.67             & 96.67                \\ \cline{2-5} 
\cellcolor[HTML]{FFFFFF}                                                                                                      & MarathiBert V2        & 97.45             & 97.45             & 97.45                \\ \cline{2-5} 
\cellcolor[HTML]{FFFFFF}                                                                                                      & MahaTweetsBERT        & 96.86             & 96.86             & 96.86                \\ \cline{2-5} 
\multirow{-4}{*}{\cellcolor[HTML]{FFFFFF}\begin{tabular}[c]{@{}c@{}}HASOC 21 + HASOC 22 + \\ MahaHate 4c-subset\end{tabular}} & MahaTweetsBERTHateful & 96.86             & 96.86             & 96.86                \\ \hline
\end{tabular}
\end{table}

\section{Results}

We hereby present our results on the HASOC 2022 dataset testing split. We report macro-F1 scores to get a clear idea of the performance of models. %Our results on the validation split of the HASOC-22 dataset are presented in Table \ref{tab:validation_results}.

We performed several ablations with MOLD v2 / HASOC 2022 shared-task test dataset \cite{hasoc2022overview}. We present these ablations in Table \ref{tab:ablation_results}. Our experiments follow the same outline: we use modified training data to train the model, and use the HASOC 2022 validation data for choosing the best checkpoint of the model. Finally, we calculate the score of each model on the MOLD v2/HASOC 2022 test set. The training datasets used for training the models are as follows:
\begin{enumerate}
    \item HASOC 2022 training data.
    \item HASOC 2022 training data and HASOC 2021 data.
    \item HASOC 2022 training data, HASOC 2021 data combined with the MahaHate 2 class dataset.
    %\item HASOC 2022 training data, HASOC 2021 data combined with the MahaHate 4 class dataset, were all examples from classes other than not offensive (NOT) were considered to be offensive (OFF) examples (MahaHate 4c-transformed).
    \item HASOC 2022 training data, HASOC 2021 data combined with the MahaHate 4 class dataset, where only offensive (OFF) and not offensive (NOT) labels are considered (MahaHate 4c-subset).

%     Hasoc 21, 22 MahaHate 2c data
% Hasoc 21, 22 MahaHate 4c (converting all non hate labels to OFF)
% Hasoc 21, 22 MahaHate 4c (dropping all labels except OFF and NOT)
\end{enumerate}

We make some key observations from the results. We also spot some interesting patterns that might help in future work.

\begin{enumerate}
    \item \textbf{Combined datasets perform considerably better than only HASOC-22 dataset:} We see that the combination of the HASOC 22 dataset along with the HASOC-21 dataset and the combination of these datasets along with the MahaHate dataset, when used to train the models, outperform the performance obtained when trained only on the HASOC-22 dataset. Although expected, this result shows that the language models have not reached their saturation point and can be scaled even further to larger data corpora.

    \item \textbf{MahaTweetBERT outperforms other models: }MahaTweetBERT is a model pre-trained on a large corpus of Marathi tweets. We observe that this model outperforms all other models in terms of macro F1. We speculate this is because the downstream dataset, the HASOC-22 dataset, has a high correlation with the pre-training dataset, which is comprised of Marathi Tweets. This shows the importance of domain-specific pre-training in NLP.
    \item \textbf{Both models have the same incorrect examples: }The model trained on the combined dataset and the one trained only on the HASOC-22 dataset incorrectly predict the same set of examples. The model trained on the smaller dataset has other incorrectly predicted examples.
    \item \textbf{Hateful BERT is not always helpful:} In general, on almost all combinations of fine-tuning data, the model pre-trained on exclusively hateful data performs worse than the model pre-trained on mixed tweets data. The MahaTweetBERT-Hateful model performs the best when only HASOC 22 data is used for training. However, as we add more datasets the benefits are not visible and the BERT trained on full tweets corpus outperforms everyone. This shows that in very low resource conditions the hateful BERT might be helpful however in general model trained on (hateful + non-hateful) corpus performs better. These observations are in line with \cite{patankar2022spread} wherein they perform a more extensive ablation.
\end{enumerate}

\section{Conclusion}
In this paper, we evaluate the performance of various BERT models on the HASOC 2022 / MOLD v2 to observe the ability of our models to detect hate speech in Marathi. We fine-tune models like MuRIL, MahaTweetBERT, and a domain-specific model, MahaTweetBERT-Hateful, pre-trained on 1 million hateful data samples on both datasets. Our experiments show that the models fine-tuned on the combined dataset perform significantly better. The MahaTweetBERT, pre-trained on 40 million Marathi tweets, outperforms all the other models. We also utilize external data sources like HASOC 2021 and MahaHate Marathi hate speech detection corpus and present an effective data augmentation strategy. We observe that models fine-tuned on both datasets fail to classify some common sentences correctly. In the future, we would like to investigate the reason for this phenomenon.

\section*{Acknowledgements}
This work was done under the L3Cube Pune mentorship
program. We would like to express our gratitude towards
our mentors at L3Cube for their continuous support and
encouragement.

\bibliography{anthology, custom}

\appendix

\end{document}